\documentclass[conference]{IEEEtran}
\IEEEoverridecommandlockouts
\usepackage{cite}
\usepackage{amsmath,amssymb,amsfonts}
\usepackage{algorithmic}
\usepackage{graphicx}
\usepackage{textcomp}
\usepackage{xcolor}
\usepackage{bm}
\usepackage{algorithm}
\def\BibTeX{{\rm B\kern-.05em{\sc i\kern-.025em b}\kern-.08em
    T\kern-.1667em\lower.7ex\hbox{E}\kern-.125emX}}

\begin{document}

\title{Diff4VS: HIV-inhibiting Molecules Generation with Classifier Guidance Diffusion for Virtual Screening
}

\author{
	\IEEEauthorblockN{
		Jiaqing Lyu\IEEEauthorrefmark{2}, 
		Changjie Chen\IEEEauthorrefmark{3}, 
        Bing Liang\IEEEauthorrefmark{4},
        Yijia Zhang\IEEEauthorrefmark{1}\IEEEauthorrefmark{2} 
		} 
  
    \IEEEauthorblockA{\IEEEauthorrefmark{2}\textit{Information Science and Technology College, Dalian Maritime University, China}, 
jiaqinglyu@hotmail.com}
    
    \IEEEauthorblockA{\IEEEauthorrefmark{3}\textit{School of Computer Science and Technology, Dalian University of Technology, China}, chenchangjie@mail.dlut.edu.cn}
    \IEEEauthorblockA{\IEEEauthorrefmark{4}\textit{School of Innovation and Entrepreneurship, Dalian University of Technology, China}, liangbing@dlut.edu.cn}

    \IEEEauthorblockA{\IEEEauthorrefmark{1}Correspondence: zhangyijia@dlmu.edu.cn}
    
} 

\maketitle

\begin{abstract}
The AIDS epidemic has killed 40 million people and caused serious global problems. The identification of new HIV-inhibiting molecules is of great importance for combating the AIDS epidemic. Here, the Classifier Guidance Diffusion model and ligand-based virtual screening strategy are combined to discover potential HIV-inhibiting molecules for the first time. We call it Diff4VS. An extra classifier is trained using the HIV molecule dataset, and the gradient of the classifier is used to guide the Diffusion to generate HIV-inhibiting molecules. Experiments show that Diff4VS can generate more candidate HIV-inhibiting molecules than other methods. Inspired by ligand-based virtual screening, a new metric DrugIndex is proposed. The DrugIndex is the ratio of the proportion of candidate drug molecules in the generated molecule to the proportion of candidate drug molecules in the training set. DrugIndex provides a new evaluation method for evolving molecular generative models from a pharmaceutical perspective. Besides, we report a new phenomenon observed when using molecule generation models for virtual screening. Compared to real molecules, the generated molecules have a lower proportion that is highly similar to known drug molecules. We call it Degradation in molecule generation. Based on the data analysis, the Degradation may result from the difficulty of generating molecules with a specific structure in the generative model. Our research contributes to the application of generative models in drug design from method,  metric, and phenomenon analysis.
\end{abstract}


\section{Introduction}
The AIDS epidemic is a grave global public health issue today, having resulted in approximately 40 million deaths. Moreover, AIDS places a significant strain on economic progress and contributes to the stigmatization of specific groups\cite{lyons2023associations}. Despite the availability of diverse anti-HIV medications, challenges such as adverse effects, drug resistance, and exorbitant treatment costs persist. Consequently, the discovery of novel HIV-inhibiting compounds stands as an urgent and imperative endeavor \cite{xu2023opportunities}.

With the development of deep learning, researchers began to try to use neural network models to generate specific content, such as images or text. VAE(Variational Autoencoders)\cite{kingma2013auto},GAN(Generative Adversarial Networks)\cite{goodfellow2014generative}, Diffusion\cite{sohl2015deep,ho2020denoising}, and other generative models have shown powerful capabilities. At the same time, several molecular generation models have been proposed, such as charRNN\cite{segler2018generating}, JT-VAE\cite{jin2018junction}, LatentGAN\cite{prykhodko2019novo}, DiGress\cite{vignac2022digress}. The application of generative models in drug design is a valuable research direction\cite{tong2021generative}.

Virtual screening is a mature computer-aided drug design technique that plays a crucial role in modern drug discovery pipelines\cite{scior2012recognizing}.  Ligand-based virtual screening methods are employed to identify novel hit candidates by retrieving compounds with high similarity to known drug molecules. A recent study combines molecular generative models with virtual screening and has made some progress\cite{kong2024discovery}.

In this paper, we introduce a pioneering approach: Classifier Guidance Diffusion for virtual screening, termed Diff4VS. This method enhances the model's capability to generate molecules that are more similar to known drugs, thereby increasing the proportion of HIV-inhibiting candidates. By integrating classifiers to guide the generative process, Diff4VS represents a significant advance in molecule generation models for virtual screening. Comparison experiments show that our method generates more HIV-inhibiting candidates than other models. Ablation experiments show that Classifier Guidance with Binary Cross Entropy(BCE) Loss is effective.

Besides, evaluating the performance of molecular generative models remains a formidable challenge\cite{gangwal2024unlocking}. Traditional metrics often fall short in capturing the nuanced requirements of drug-likeness, which is paramount in the practical application of generated molecules. Inspired by ligand-based virtual screening, a new metric DrugIndex is proposed. The DrugIndex is the ratio of the proportion of candidate drug molecules in the generated molecule to the proportion of candidate drug molecules in the training set. This metric is crucial for ensuring that the generated molecules possess characteristics that are conducive to drug design.

Additionally, through our experiments, we report and discuss a new phenomenon observed during virtual screening processes. We found that molecules generated by current molecule generation models exhibit a lower proportion of high similarity to known drug molecules compared to real molecules. We call it Degradation in molecule generation. Based on the data analysis, the Degradation may result from the difficulty of generating molecules with a specific structure in the generative model.

\begin{figure*}[htbp]
\centerline{\includegraphics[width=1\linewidth]{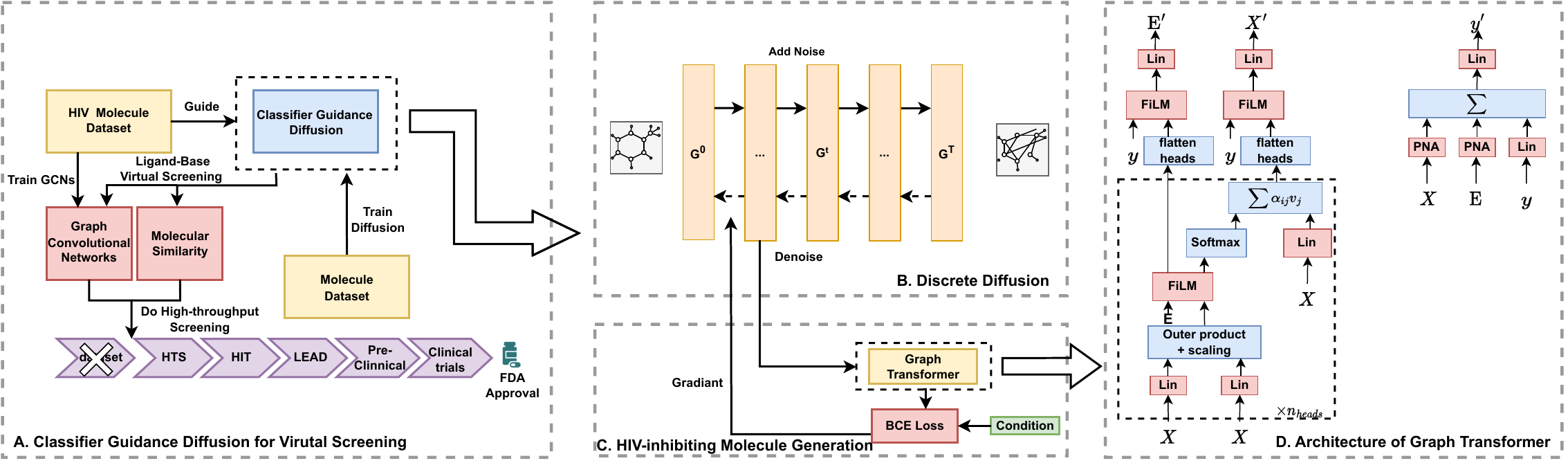}}
\caption{The Overview of Diff4VS.}
\label{fig: overview}
\end{figure*}

In summary, the contributions of this paper are summarized as follows:
\begin{itemize}

\item \textbf{Introduction of Diff4VS:} We are the first to use a Classifier Guidance Diffusion for virtual screening(Diff4VS), which enables the model to generate a larger proportion of HIV-inhibiting candidate molecules.

\item \textbf{Novel Metric:} We propose a new metric DrugIndex to evaluate the ability of molecule generation models to generate drug-like molecules for virtual screening. DrugIndex provides a new evaluation method for evolving molecular generative models from a pharmaceutical perspective.

\item \textbf{New Phenomenon:} We report and discuss a new phenomenon, Degradation. Compared to real molecules, the generated molecules have a lower proportion that is highly similar to known drug molecules. 

\end{itemize}

\section{Related Work}
\label{sec:aigc}
\subsection{Variational Autoencoders for Molecule Generation}
\label{sec:vae}
The core idea of Variational Autoencoders(VAE)\cite{kingma2013auto} is to map the input data to latent variables in the latent space via an encoder and the latent variables back to the original data space via a decoder. Both encoders and decoders are neural network models trained to maximize the similarity between the original data and the reconstructed data. Researchers have proposed several VAE models that can generate molecules, such as JT-VAE\cite{jin2018junction}. Conditional Variational Autoencoder(cVAE)\cite{sohn2015learning} is an extended model based on VAE to generate samples with conditions.

\subsection{Generative Adversarial Networks for Molecule Generation}
\label{sec:gan}
The working principle of Generative Adversarial Networks(GAN)\cite{goodfellow2014generative} is based on the idea of adversarial training from game theory. The generator receives random noise as input and tries to generate samples similar to the real data. The classifier, in turn, receives both real samples and samples generated by the generator and tries to distinguish their sources. During the training process, the generator gradually improves the quality of the generated samples. At the same time, the classifier improves the accuracy of discrimination by continuously distinguishing between real data and generated data. Researchers have proposed several GAN models that can generate molecules, such as LatentGAN\cite{prykhodko2019novo}. Conditional Generative Adversarial Networks (cGAN) is a variant of GAN.

\subsection{Diffusion Models For Molecule Generation}
\label{sec:related_work_diffusion}
The Diffusion model\cite{sohl2015deep,ho2020denoising} constructs two parameterized Markov chains to diffuse the data with predefined noise and reconstruct the desired samples from the noise. In the forward chain, the Diffusion model gradually adds pre-designed noise to real samples until the samples conform to a specific distribution, such as a Gaussian distribution. At the same time, real samples and noisy samples are used to train a neural network to predict the distribution of noise. Starting from the specific distribution, the reverse chain uses a trained neural network to sample noise from the predicted distribution and remove the noise from samples. Researchers have proposed several Diffusion models that can generate molecules, such as DiGress\cite{vignac2022digress}.

\section{Method}
In this section, we provide a detailed introduction to our proposed method, Diff4VS. As shown in Fig.~\ref{fig: overview}, our method mainly consists of four parts. We introduce them in Section~\ref{sec:Diff4VS}, Section~\ref{sec:forward}, Section~\ref{sec:condition} and Section~\ref{sec:architecture}. Then we introduce our new metric in Section~\ref{sec:metric}.

\subsection{Diffusion for Virtual Screening}
\label{sec:Diff4VS}
Ligand-based virtual screening detects molecules similar to known drugs from datasets. However, it is limited by molecular datasets. If there is a lack of molecules similar to known drugs in the dataset, virtual screening cannot find candidates. Therefore, the use of known drug molecules to guide generative models to generate candidates for virtual screening is noteworthy.

We are the first to combine conditional Diffusion with virtual screening. This new paradigm is shown in Fig.~\ref{fig: overview}-A. First, with millions of molecules, the Diffusion model is trained to generate molecules. Second, with molecules in HIV dataset, an extra classifier is trained to guide Diffusion for HIV-inhibiting molecule generation. Third, ligand-base virtual screening technologies are used to detect HIV-inhibiting candidates from generated molecules.

\subsection{Discrete Diffusion}
\label{sec:forward}
A molecule can be viewed as a graph $G$, atoms are vertexes and bonds are edges. Inspired by Austin et al\cite{austin2021structured}, one-hot encoding is used to represent the class of atoms and bonds. Noise is added separately on each atom and bond feature $X$. Instead of adding Gaussian noise, the feature of each atom and bond are multiplied by a matrix $Q$. $[Q]_{ij}^{t}$ represents the probability of jumping from state $i$ to state $j$ at round $t$, and $\overline{Q}^t=\prod_{i=1}^{t}Q^i$. Eq~\ref{eq:add_noise_discrete} is used to add noise. 

\begin{equation}
\label{eq:add_noise_discrete}
q(X_t|X_{t-1}) = X_{t-1}Q^t \quad and \quad q(X_t|X_0)=X_0\overline{Q}^t
\end{equation}

Inspired by Vignac et al\cite{vignac2022digress}, the transition matrices definition for atoms and bonds are from the marginal distribution. Note that the bonds of molecules are undirected, only the upper-triangular part of $B$ is added with noise and then the $B$ will be symmetrized.

In the reverse process, $q(x^{t-1}|G^t)$ is calculated for denoising. There is a limited variety of atoms and bonds. So the kinds of atoms and bonds can be enumerated. $\mathbb{X}$ is the set of kinds. For feature $x^{t-1}$, $q(x^{t-1}|G^t)$ can be calculated with Eq~\ref{eq:x_given_G}. 

\begin{equation}
\label{eq:x_given_G}
\begin{aligned}
q(x^{t-1}|G^t)=\sum_{x \in \mathbb{X}} q(x^{t-1}|x^0=x,x^t)p(x^0=x|G^t)
\end{aligned}
\end{equation}

Vignac et al\cite{vignac2022digress} proved that $q(x^{t-1}|x^0=x,x^t)$  is proportional to $ x^t(Q^t)^\top \odot x^0\overline{Q}^{t-1}$. According to Yang et al\cite{song2019generative}, when $q(x^{t-1}|x^t,x^0)$ has a closed-form expression, $x_0$ can be used as the target of the neural network. 

As a result, a neural network $f_\theta(G^t,t)$ is trained to predict $q_\theta(G^0|G^t)$ in the forward process of discrete Diffusion. With $p_\theta(x^0=x|G^t)$, $q_\theta(x^{t-1}|G^t)$ can be calculated. 

Inspired by \cite{satorras2021n}, a loss function with invariance is used in Eq~\ref{eq:loss}. $\hat{p}^{G^0}$ is the predicted probabilities for each atom and bond in $G^0$. $\lambda$ controls the relative importance of nodes and edges, and we set it to $5.0$ in our experiments. The training process of the discrete Diffusion model is shown in Algorithm~\ref{al:train_diffuison}.
\begin{algorithm} 
	\caption{The training process of the discrete Diffusion} 
	\label{al:train_diffuison} 
	\begin{algorithmic}
		\REQUIRE $G=(\bm{A},\bm{B})$ and $T$
		\STATE Sample $t$ from $\mathcal{U}(1,T)$  
		\STATE Sample $G^t$ from $q(G^t|G^0)$
		\STATE $z \gets$ merge\_t\_cycle\_spectral\_feature($G^t,t$)
            \STATE $q_\theta(G^0|G^t) \gets f_\theta(G^t,z)$
            \STATE $l \gets$ $loss(q_\theta(\bm{A}^0|G^t),\bm{A})+\lambda loss(q_\theta(\bm{B}^0|G^t),\bm{B})$
            
            \STATE optimizer.step($l$)
	\end{algorithmic} 
\end{algorithm}

\begin{equation}
\label{eq:loss}
l(\hat{p}^{G^0},G^0)=\sum_{1\leq i \leq n}\mathrm{CE}(\hat{p}_i^{A^0}, a_i^0) + \lambda\sum_{1\leq j \leq n^2}\mathrm{CE}(\hat{p}_j^{B^0}, b_j^0) 
\end{equation}

\subsection{HIV-inhibiting Molecule Generation}
\label{sec:condition}
We take HIV-inhibiting molecule generation as condition $y_c$. The value of $y_c$ is either 0 or 1. Since $G^t$ and $y_c$ are given, and $\hat{p}(y_c|G^t)$ is independent of $G^{t-1}$, $1/\hat{p}(y_c|G^t)$ is a constant $Z$. As a result, $\hat{q}(G^{t-1}|G^t,y_c)$ is shown in Eq~\ref{eq:conditional_ddpm}.

\begin{equation}
\label{eq:conditional_ddpm}
\begin{aligned}
\hat{q}(G^{t-1}|G^t,y_c) = \hat{q}(G^{t-1}|G^t)\hat{q}(y_c|G^{t-1},G^t)Z
\end{aligned}
\end{equation}

The noise-adding process is the same whether it is a conditional generation or not. As a result,$\hat{q}(G^{t-1}|G^t)$ is equal to $q(G^{t-1}|G^t)$. And $\hat{q}(G^{t-1}|G^t,y_c)$ can be calculated with $q(G^{t-1}|G^t)\hat{q}(y_c|G^{t-1})Z$.

We show the way to calculate $q(G^{t-1}|G^t)$ in Section~\ref{sec:forward}, and $\hat{q}(y_c|G^{t-1})$ is equivalent to the result of a classifier with noise. That's why Classifier-Guidance Diffusion trains a classifier $g_\theta$ with noise. However, there are so many possible values of $G^{t-1}$(Atoms:$8^n$ $\times$ Bonds: $4^{(n*n)}$). As a result, $\hat{q}(y_c|G^{t-1})$ cannot be obtained directly from the classifier $g_\theta$. 

Inspired by Dhariwal et al\cite{dhariwal2021diffusion} and Vignac et al\cite{vignac2022digress}, we view $G$ as a continuous tensor of order $n + n^2$($n$ atoms and $n^2$ bonds). Using the first-order Taylor expansion around $G^t$, $C$ is a constant independent of $G^{t-1}$, and $\hat{q}(y_c|G^{t-1})$  is approximately equal to Eq~\ref{eq:Taylor}.

\begin{equation}
\label{eq:Taylor}
\begin{aligned}
\log \hat{q}(y_c|G^{t-1}) &\approx \log  \hat{q}(y_c|G^t) + ⟨\nabla_G,G^{t-1}-G^t⟩\\
&= ⟨\nabla_G,G^{t-1}⟩ + C
\end{aligned}
\end{equation}
 
We make the additional assumption that $\nabla_G\log\hat{q}(y_c|G^t)$ is proportional to $-\nabla_G(y_c\log y_\theta + (1-y_c)\log(1 - y_\theta))$. Therefore, we use the Binary Cross Entropy(BCE) as the loss function for the classifier $g_\theta$ and then estimate $\nabla_G\log\hat{q}(y_c|G^t)$ with the gradient of the classifier $g_\theta$. So we can calculate $\hat{q}(y_c|G^{t-1},G^t)$ and $\hat{q}(G^{t-1}|G^t,y_c)$. Based on $\hat{q}(G^{t-1}|G^t,y_c)$, we sample $G^{t-1}$. This process is repeated and finally, $G^0$ is generated. The sampling process of the discrete Diffusion mode is shown in Algorithm~\ref{al:sample_diffusion}.

\begin{algorithm} 
	\caption{The sampling process of the discrete Diffusion} 
	\label{al:sample_diffusion} 
	\begin{algorithmic}
		\REQUIRE $T$
		\STATE Sample $G^T$ from marginal distribution
           
    \FOR{$i=T$ to $1$}
        \STATE $z \gets$ merge\_t\_cycle\_spectral\_feature($G^t,t$)
        \STATE $q_\theta(G^0|G^t) \gets f_\theta(G^t,z)$
        \STATE $q(G^{t-1}|G^t,G^0) \propto G^t(Q^t)^\top \odot G^0\overline{Q}^{t-1}$
        \STATE $q(a_i^{t-1}|G^t)=\sum_{a \in \mathbb{A}} q(a_i^{t-1}|a_i^0=a,a_i^t)p(a_i^0=a|G^t)$
        \STATE $q(b_i^{t-1}|G^t)=\sum_{b \in \mathbb{B}} q(b_i^{t-1}|b_i^0=b,b_i^t)p(b_i^0=b|G^t)$
        \STATE $y_\theta \gets g_\theta(G^t,t)$
        \STATE $\hat{q}(y_c|G^{t-1}) \propto \exp(-\lambda⟨\nabla_G\log\hat{q}(y_c|G^t),G^{t-1}⟩ )$
        \STATE $\hat{q}(G^{t-1}|G^t,y_c) \gets \hat{q}(G^{t-1}|G^t)\hat{q}(y_c|G^{t-1})$
        \STATE Sample $G^{t-1}$ from $\hat{q}(G^{t-1}|G^t,y_c)$
    \ENDFOR
    
    \end{algorithmic} 
\end{algorithm}

\subsection{Network Architecture}
\label{sec:architecture}
A neural network $f_\theta(G^t,t)$ is trained to predict $q_\theta(G^0|G^t)$ in the forward process of discrete Diffusion.  Inspired by Vignac et al\cite{vignac2022digress}, cycles and spectral features are added as extra inputs to improve performance. We use graph transformer network\cite{dwivedi2021generalization} with FiLM layers\cite{perez2018film} to predict $q_\theta(G^0|G^t)$. Besides, we use the same graph transformer for classifier guidance. The network architecture is shown in Fig~\ref{fig: overview}-D.  

\subsection{Proposed Metric}\label{sec:metric}
Molecule generation models have found one important application in the field of drug molecule discovery. Table~\ref{tab: metrics} outlines several common evaluation metrics used to assess the performance of generative models. However, beyond QED, these evaluation metrics do not specifically focus on the ability of the models to generate drug-like molecules. Therefore, the existing metrics are unable to meet the requirements of drug molecule design with generative models.

\begin{table}[htbp]
  \centering
  \caption{Some Evaluation Metrics for Molecule Generation Models}
    \begin{tabular}{ccc}
    \hline
    Metrics & Class & Description \\
    \hline
    Valid\cite{brown2019guacamol} & Prop. & Molecules that pass basic valency checks \\
    Unique\cite{brown2019guacamol} & Prop. & Molecules that have different SMILES strings \\
    Novel\cite{brown2019guacamol} & Prop. & Molecules that are not in the training set \\
    Filters\cite{vignac2022digress} & Prop. & Molecules that pass the filters for building set \\
    QED\cite{vignac2022digress} & Avg. & Quantitative estimation of drug-likeness \\
    SA\cite{ertl2009estimation}    & Avg. & Synthetic accessibility score \\
    \hline
    \end{tabular}%
  \label{tab: metrics}%
\end{table}%

Inspired by ligand-based virtual screening approaches, we design a novel evaluation metric to better assess the capacity of models to generate drug-like molecules. $f_a$ denotes the morgan fingerprint of the molecule $a$. The similarity between molecule $a$ and $b$ is defined in Eq.~\ref{eq: molsim}. The similarity of a molecule $b$ to a set of molecules $A$ is defined in Eq.~\ref{eq: setmolsim}. Besides, Eq.~\ref{eq: druglike} calculates the proportion of molecules in set $B$ that are similar to molecules in the known drug molecule set $A$.

\begin{equation}
MolSim(a,b)=TanimotoSimilarity(f_a,f_b)
\label{eq: molsim}
\end{equation}

\begin{equation}
SetMolSim(A,b)=\mathop{\arg\max}\limits_{a \in A}MolSim(a,b)
\label{eq: setmolsim}
\end{equation}

\begin{equation}
DrugLike(A,B)=\frac{\{b\in B|SetMolSim(A,b) > 0.5\}}{|B|}
\label{eq: druglike}
\end{equation}

The set of known drug molecules is denoted as $X$, the set of molecules used to train the generative model is denoted as $Y$, and the set of molecules generated by the generative model is denoted as $Z$. The new metric we define is as shown in Eq.~\ref{eq: collapse}. We refer to it as the Drug Index. The Drug Index reflects the ability of generative models to generate drug-like molecules. 

\begin{equation}
DrugIndex(X,Y,Z)=\frac{DrugLike(X,Z)}{DrugLike(X,Y)}
\label{eq: collapse}
\end{equation}

The Drug Index is inspired by virtual screening and is designed to be intuitive and easy to implement. Using the Drug Index, one can evaluate molecule generation models that are suitable for computer-aided drug design tasks such as virtual screening.

\section{Experiments}
First, we compare our method with others from Sections~\ref{sec:dataset} through~\ref{lab:gcn_comparison}. Then we conduct ablation experiments in Section~\ref{sec:ablation} and present some generated candidate molecules in Section~\ref{sec:casestudy}. Additionally, we report and discuss the Degradation phenomenon in Section~\ref{sec:phen}. Finally, we discuss the limitations of our work in Section~\ref{sec:limitation}. All the code and results will be made available on Github.\footnote{https://github.com/jiaqinglv2000/Diff4VS}.

\begin{figure*}[htbp]
\centerline{\includegraphics[width=1\linewidth]{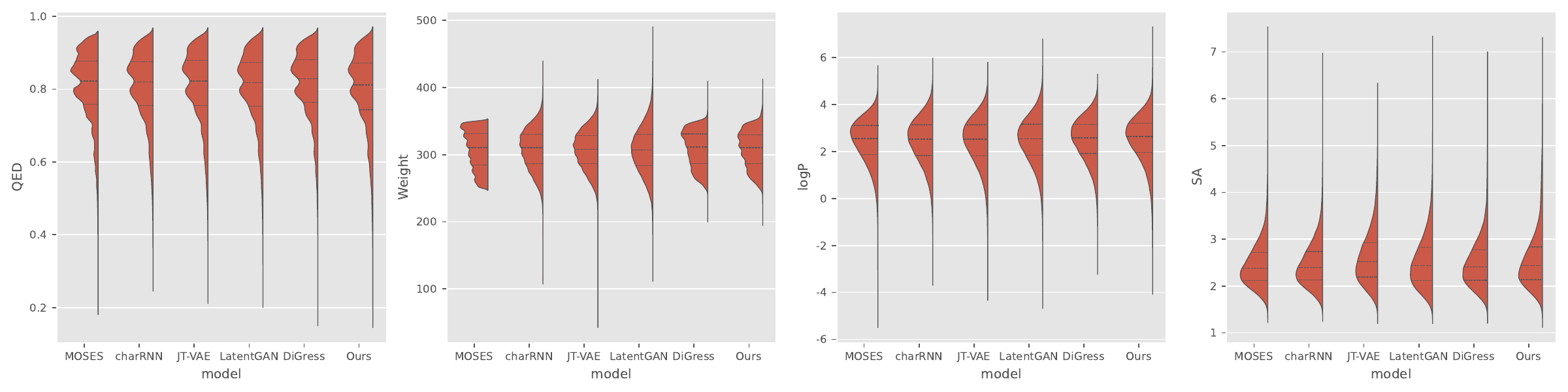}}
\caption{The property distribution of molecules from MOSES and generative models.}
\label{fig:property}
\end{figure*}

\subsection{Dataset}
\label{sec:dataset}
All molecule generation models are trained on the MOSES\cite{polykovskiy2020molecular} dataset and its properties are listed in Table ~\ref{tab:moses}.

\begin{table}[htbp]
  \centering
  \caption{The properties of the MOSES dataset}
    \begin{tabular}{cc}
    \hline
    Properties & Values \\
    \hline
    Molecular weight(Dalton) & 250-350 \\
    Number of rotatable bonds & $\leq$ 7 \\
    XlogP & $\leq$ 3.5 \\
    Molecules containing charged atoms & Removed \\
    Atoms & C, N, S, O, F, Cl, Br, H \\
    Bonds & single, double, triple, aromatic \\
    Cycles & No cycles larger than 8 atoms \\
    Train dataset size & 1584664 \\
    Valid dataset size & 176075 \\
    Test  dataset size & 176226 \\
    \hline
    \end{tabular}%
  \label{tab:moses}%
\end{table}%

The HIV dataset is a part of the MoleculeNet\cite{wu2018moleculenet} dataset which tested the ability to inhibit HIV replication. There are 41127 molecules in the HIV dataset. Each molecule has an active or inactive label indicating whether it inhibits HIV replication. Referring to the restrictions of molecular weight, atom type, and bond type in Table~\ref{tab:moses}, we construct a subset HIV-a. As shown in Table~\ref{tab:hiv}, the HIV-a dataset contains 12116 molecules.

We perform simple upsampling on the HIV-a dataset, replicating the positive samples until the number of positive and negative samples is roughly balanced. The classifier $g_\theta$ of Diff4VS is trained on the HIV-a dataset with noise for guidance. The hyperparameter $\lambda$ is set to 1000.

In addition to these 12,116 molecules, the remaining 29,011 molecules were used to train Graph Neural Networks (GNNs) for evaluating the generated molecules, as detailed in Section~\ref{sec:gnns}. We refer to these 29,011 molecules as the HIV-b dataset.

\begin{table}[htbp]
  \centering
  \caption{The details of the HIV-a dataset}
    \begin{tabular}{cccc}
    \hline
          & Train & Valid & Test \\
    \hline
    Active & 328   & 30    & 33 \\
    Inactive & 9672  & 1086  & 967 \\
    \hline
    \end{tabular}%
  \label{tab:hiv}%
\end{table}%

\subsection{Comparison across Generative Models}
\label{sec:comparisonofDI}
We generate 90,000 molecules using charRNN\cite{segler2018generating}, JT-VAE\cite{jin2018junction}, LatentGAN\cite{prykhodko2019novo}, DiGress\cite{vignac2022digress}, and our method. We used the known HIV inhibitor molecules from the MoleculeNet dataset \cite{wu2018moleculenet} as the reference drug molecules. The HIV Drug Index (HIV DI), Quantitative Estimate of Druglikeness (QED), and other properties of these generative models are shown in Table~\ref{tab: hivDrugIndex}.

\begin{table}[htbp]
  \centering
  \caption{Comparison of Drug Index across Generative Models}
    \begin{tabular}{cccccc}
    \hline
          & charRNN & JT-VAE & LatentGAN & DiGress & Ours \\
    \hline
    Arch. & RNN   & VAE   & GAN   & Diff. & Diff. \\
    Class & SMILES & Tree  & SMILES & Graph & Graph \\
    Valid & 87730 & 89998 & 80698 & 78105 & 71056 \\
    Weight & 308.23 & 306.68 & 306.83 & 308.16 & 307.82 \\
    logP & 2.44 & 2.42 & 2.47 & 2.49 & 2.52 \\
    SA & 2.47 & 2.61 & 2.52 & 2.49 & 2.56  \\
    QED   & 0.804 & 0.805 & 0.802 & 0.811 & 0.795 \\
    HIV DI & 78.80\% & 87.46\% & 94.19\% & 86.11\% & $\bm{104.62\%}$ \\
    \hline
    \end{tabular}%
  \label{tab: hivDrugIndex}%
\end{table}%

According to Table~\ref{tab: hivDrugIndex}, the HIV Drug Index provides a more effective way to compare the ability of different generative models to generate HIV drug molecules. Unlike the HIV Drug Index, the difference in the average QED values of molecules generated by different generative models is often negligible, making it challenging to discern differences in their ability to generate drug-like molecules. In contrast, the newly designed Drug Index metric exhibits certain advantages over using QED alone.

Besides, our method is the only one with a Drug Index greater than $100\%$. The Drug Index of our method is $10.43\%$ higher than the second-best method. This demonstrates that our method can generate more molecules that are structurally similar to existing HIV drugs. Furthermore, virtual screening with our method can identify more candidate drug molecules.

According to the information in Table~\ref{tab: hivDrugIndex} and Fig.~\ref{fig:property}, our method of generating molecules has similar properties such as Weigh, logP, and SA compared to other generation methods, and the distribution is close to that of MOSES. This indicates that the guidance of our classifier has not disrupted the distribution of the original generative model. The average QED of our method is the lowest, it is because the average QED of active molecules in the HIV dataset is also lower than MOSES(0.40 vs 0.81). Therefore, the lowest QED does not negate the ability of our method to generate anti-HIV drug candidates.

\begin{figure*}[htbp]
\centerline{\includegraphics[width=1\linewidth]{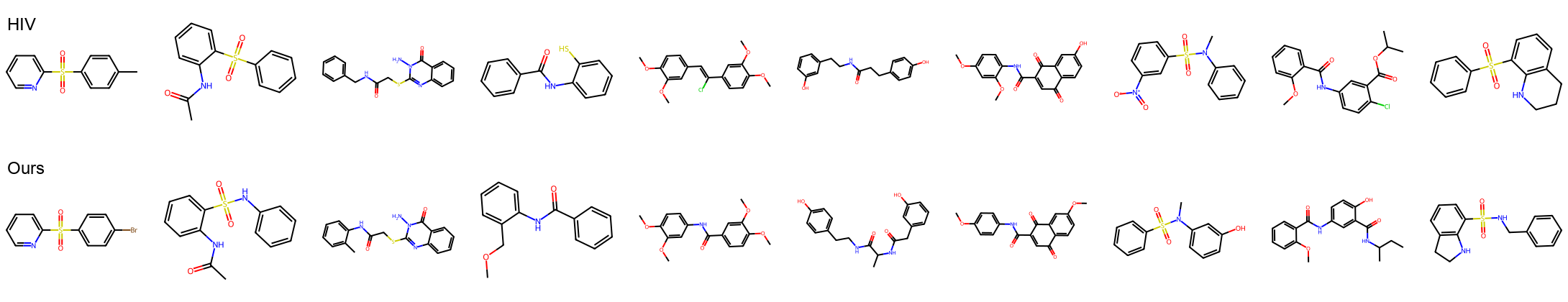}}
\caption{Pairs of similar molecules from our model and HIV dataset.}
\label{fig:mol_pair}
\end{figure*}

\subsection{Comparison of GNNs for Virtual Screening}
\label{sec:gnns}
Neural networks are also a way of ligand-based virtual screening. We select four representative GNNs and train them on the HIV-b dataset. The four GNNS are GAT(Graph Attention Network)\cite{velickovic2017graph}, GCN(Graph Convolution Network)\cite{kipf2016semi}, MPNN(Message Passing Neural Network)\cite{gilmer2017neural} and AttentiveFP\cite{xiong2019pushing}. The learning rate of the four models is set to 0.001, and the number of epochs is set to 500. Models are implemented by DeepChem\cite{Ramsundar-et-al-2019}. We do simple upsampling on the HIV-b dataset, replicating the active samples until there are roughly the same number of active and inactive samples.

To objectively reflect the ability of each model to identify active molecules, we conducted 5-fold cross-validation experiments. When the probability given by the model is greater than $50\%$, the input molecule is considered to be active. The results are shown in Table~\ref{tab:vs_performance}. There are significantly more inactive samples than active samples in the HIV-b test set. Therefore, even if the F1-score is lower than 0.5, the model still demonstrates a certain level of discriminative ability. When the ROC-AUC is larger than 0.7, the virtual screening performance is considered fair, as suggested by prior studies\cite{hamza2012ligand}. The ROC-AUC results also indicate that graph neural networks (GNNs) can be a viable approach for the virtual screening of HIV drug molecules.

\begin{table}[htbp]
  \centering
  \caption{Performance Comparision of Classifiers}
    \begin{tabular}{ccccccccc}
    \hline
          & \multicolumn{2}{c}{Precision} & \multicolumn{2}{c}{F1-value} & \multicolumn{2}{c}{ROC-AUC} \\
          & mean  & std    & mean  & std   & mean  & std \\
    \hline
    GCN   & 0.4878 & 0.0392  & \textbf{0.4772} & 0.0364 & 0.7246 & 0.0209 \\
    GAT   & 0.2980 & 0.0489 &  0.3764 & 0.0302 & \textbf{0.7374} & 0.0141 \\
    AttentiveFP & 0.4892 & 0.0594 & 0.4646 & 0.0327 & 0.7126 & 0.0085 \\
    MPNN  & \textbf{0.4994} & 0.0266 & 0.4746 & 0.0211 & 0.7172 & 0.0101 \\
    \hline
    \end{tabular}%
  \label{tab:vs_performance}%
\end{table}%

For both F1-score and ROC-AUC, GCN achieves better results than AttentiveFP and MPNN. Although GAT achieves the highest ROC-AUC, it has a low Precision. As a result, We use GCN for further verification in Section~\ref{lab:gcn_comparison}.

\subsection{Comparison of Drug Index using GCNs}
\label{lab:gcn_comparison}
To further verify the effectiveness of the our proposed metric and method, we use GCNs for additional verification. We train three GCNs on the HIV-b dataset. When the sum of the output probabilities of the three models is greater than 100\%(i.e., 2 out of the 3 models consider the molecule to be active), the input molecule is deemed active. Instead of Eq.~\ref{eq: setmolsim}, the output of GCNs is used to calculate DrugLike and DrugIndex. The results are presented in Table~\ref{tab:GCN_DI_performance}.

\begin{table}[htbp]
  \centering
  \caption{Comparison of Drug Index using GCNs}
    \begin{tabular}{cccccc}
    \hline
          & charRNN & JT-VAE & LatentGAN & DiGress & Ours \\
    \hline
    GCN DI & 96.23\% & 89.90\% & 91.35\% & 96.15\% & $\bm{112.02\%}$ \\
    \hline
    \end{tabular}%
  \label{tab:GCN_DI_performance}%
\end{table}%

Our method is the only one with a GCN-based Drug Index greater than $100\%$. The Drug Index of our method is $15.87\%$ higher than the second-best approach. These results are consistent with those presented in Section~\ref{sec:comparisonofDI}. This further demonstrates that virtual screening with our method can identify more potential HIV-inhibiting candidates.

\subsection{Ablation Experiment}
\label{sec:ablation}
To robustly validate the efficacy of our approach, we conducted ablation studies. Our primary focus was to investigate the impact of the classifier guidance in the diffusion model and our innovative adoption of the BCE loss function on the final results. We established three models: the first excluded the classifier guidance module entirely, the second employed MSE as the classifier's loss function, and the third utilized BCE as the classifier's loss function. 

The results are shown in Table~\ref{tab:Ablation}, we observed that the HIV Drug Index (HIV DI) for the model without classifier guidance and the model employing MSE as the classifier's loss function stood at $86.11\%$ and $83.51\%$ respectively. Strikingly, the HIV DI for the model utilizing BCE as the classifier's loss function soared to $104.62\%$, significantly outperforming the other two. This underscores the crucial role of the choice of loss function for classifier guidance, as an inappropriate selection can lead to diminished rather than enhanced performance. Furthermore, it solidifies the efficacy of our innovative adoption of the BCE loss function, reaffirming that our approach for virtual screening is capable of yielding a more extensive repertoire of HIV-inhibiting candidate molecules.

\begin{table}[htbp]
  \centering
  \caption{Ablation Experiment}
    \begin{tabular}{cccc}
    \hline
          & Model 1 & Model 2 & Model 3 \\
    \hline
    Classifier Guidance & $\times$    & \checkmark   & \checkmark \\
    Loss Function & -  & MSE & BCE \\
    
    HIV DrugIndex & $86.11\%$ & $83.51\%$ & $104.62\%$ \\
    \hline
    \end{tabular}%
  \label{tab:Ablation}%
\end{table}%

\subsection{Generated Candidate Molecules}
\label{sec:casestudy}
Ten pairs of similar molecules are shown in Fig.~\ref{fig:mol_pair} with the top row of active molecules from the HIV dataset and the bottom row of molecules generated by our model. Our model generates molecules that are very similar to known HIV drug molecules. This further demonstrates the effectiveness of our method.

\subsection{Observed Phenomenon "Degradation" and Discussion}\
\label{sec:phen}
In Table~\ref{tab:vs_performance} and Table~\ref{tab:GCN_DI_performance}, it can be observed that the Drug Index of the generative models(except our method) are all below $100\%$. Compared with the molecules in training set MOSES, the proportion of generated molecules that are similar to known HIV drugs is lower. We are the first to observe and report this phenomenon. We call it the \textbf{Degradation} in molecule generation.

To explain this phenomenon, we do the following experiment. First, we calculate Morgan fingerprint of all molecules with a radius of 10. Then, we use the k-means algorithm to cluster the active molecules in the HIV-a dataset according to fingerprint similarity. The $k$ is set to $30$. In the end, we use the k-nearest neighbor(KNN) algorithm to classify molecules that are similar to known drugs. These molecules are from the MOSES dataset as well as generated by  charRNN\cite{segler2018generating}, JT-VAE\cite{jin2018junction}, LatentGAN\cite{prykhodko2019novo}, DiGress\cite{vignac2022digress}, and our method. The result is shown in Table\ref{tab:knn}.

\begin{table}[htbp]
  \centering
  \caption{The Clusters of Molecules}
    \begin{tabular}{cccccccc}
    \hline
          Cluster & HIV   & MOSES & \cite{segler2018generating} & \cite{prykhodko2019novo} & \cite{vignac2022digress} & \cite{jin2018junction} & Ours \\
    \hline
    1     & 10    & 0     & 0     & 0     & 0     & 0     & 0 \\
    2     & 3     & 0     & 0     & 0     & 0     & 0     & 0 \\
    3     & 10    & 0     & 0     & 0     & 0     & 0     & 0 \\
    4     & 4     & 0     & 0     & 0     & 0     & 0     & 0 \\
    5     & 84    & 213   & 187   & 204   & 180   & 250   & 202 \\
    6     & 5     & 0     & 0     & 0     & 0     & 0     & 0 \\
    7     & 7     & 0     & 0     & 0     & 0     & 0     & 0 \\
    8     & 2     & 0     & 3     & 4     & 2     & 0     & 2 \\
    9     & 1     & 0     & 0     & 0     & 0     & 0     & 0 \\
    10    & 3     & 0     & 0     & 0     & 0     & 0     & 0 \\
    11    & 11    & 0     & 0     & 0     & 0     & 0     & 0 \\
    12    & 14    & 0     & 0     & 0     & 0     & 0     & 0 \\
    13    & 5     & 1     & 0     & 0     & 0     & 0     & 0 \\
    14    & 5     & 0     & 0     & 0     & 0     & 0     & 0 \\
    15    & 3     & 0     & 0     & 0     & 0     & 0     & 0 \\
    16    & 12    & 0     & 1     & 0     & 0     & 0     & 0 \\
    17    & 4     & 0     & 0     & 0     & 0     & 1     & 0 \\
    18    & 10    & 0     & 0     & 0     & 0     & 0     & 0 \\
    19    & 2     & 0     & 0     & 0     & 0     & 0     & 0 \\
    20    & 8     & 2     & 2     & 1     & 2     & 2     & 1 \\
    21    & 7     & 0     & 2     & 2     & 1     & 0     & 2 \\
    22    & 3     & 0     & 0     & 0     & 0     & 0     & 1 \\
    23    & 4     & 2     & 1     & 0     & 0     & 1     & 0 \\
    24    & 27    & 4     & 3     & 4     & 4     & 3     & 0 \\
    25    & 1     & 0     & 0     & 0     & 0     & 0     & 0 \\
    26    & 95    & 86    & 65    & 52    & 67    & 52    & 84 \\
    27    & 21    & 1     & 0     & 0     & 0     & 0     & 2 \\
    28    & 2     & 0     & 0     & 0     & 0     & 0     & 0 \\
    29    & 1     & 0     & 0     & 0     & 0     & 0     & 0 \\
    30    & 27    & 59    & 68    & 98    & 67    & 69    & 63 \\
    \hline
    \end{tabular}%
  \label{tab:knn}%
\end{table}%

According to Table~\ref{tab:knn}, the following conclusions can be inferred.

\subsubsection{Degradation from Difficulty in Specific Structures}
According to cluster 13, we can find that there are only molecules from MOSES in this cluster and no molecules from the generative models. Although the training set contains similar molecules, the generative model has difficulty generating the structure of this cluster of HIV drug molecules. Therefore, we can speculate that the inability to generate these specific structures is the reason for the poor performance. Actually, some scholars have found that generation models often struggle to generate aromatic structures\cite{vignac2022digress}, which is consistent with our conclusion.

\subsubsection{Training Set Needs Improvement}
For most clusters, there are neither molecules from the MOSES dataset nor molecules generated by the existing generative models. The molecules in the MOSES dataset lack certain molecular structures, so the generation models trained on this dataset struggled to produce molecules with these specific structural features. As a result, improving the training data for generative models is an important consideration in the drug discovery process.

\subsubsection{Our Method Makes Sense}
Our method uniquely generates candidate molecules within the cluster 22. This demonstrates the effectiveness of our approach in leveraging existing HIV-inhibiting molecules to guide the generation model.

\begin{figure}
    \centering
    \includegraphics[width=0.10\textwidth]{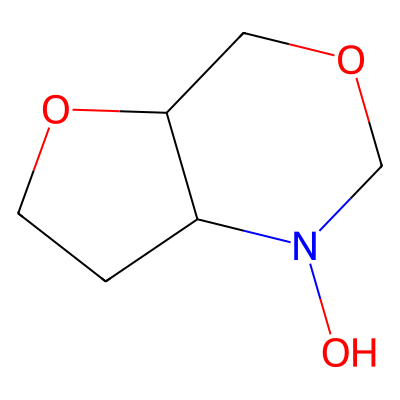}
    \caption{Certain molecular structure.}
    \label{fig:56}
\end{figure}

To further validate our conclusion, we find that many clusters containing HIV molecules without any molecules from MOSES or the generative models possessed a similar structure to Fig.~\ref{fig:56}, where two rings of length 5 or 6 share at least one chemical bond with each other. 

\begin{figure}
    \centering
    \includegraphics[width=0.4\textwidth]{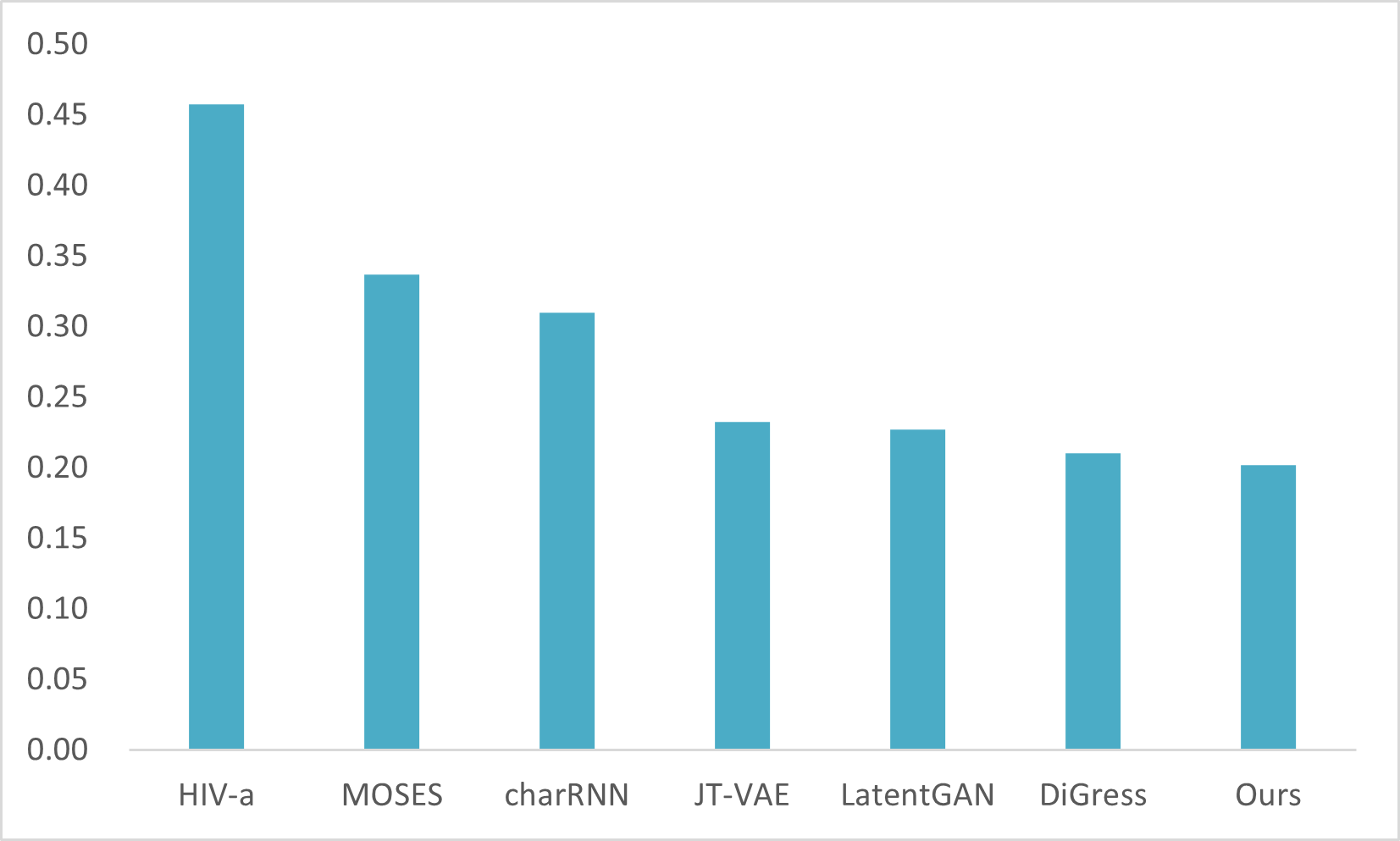}
    \caption{The proportion of molecules containing the particular structure.}
    \label{fig:56tab}
\end{figure}

We have compiled the proportion of molecules with this structure in candidate molecule datasets, as shown in Fig~\ref{fig:56tab}. It is evident that the proportion of molecules with this structure in the generative models' datasets is lower than that in MOSES, while the proportion in MOSES is still lower than that in HIV. This further demonstrates the difficulty of existing generative models in generating specific structures and the need for improvement in the training set. Some previous studies have shown that GNN is difficult to handle rings\cite{Loukas2020What}. As a result, SMILES-based charRNN generates the most candidate molecules with this structure than other models.

Although our method generated fewer candidate molecules with this particular structure, we believe that Diff4VS has advantages in generating other drug molecule structures, given the overall advantages of our method in the DrugIndex.

\subsection{Limitation}
\label{sec:limitation}
\subsubsection{Lack of Data}
The training set for the Classifier Guidance contains only 328 HIV-inhibiting molecules. Although Classifier Guidance Diffusion requires less data than Classifier-Free Diffusion, it is still challenging to train a reliable classifier for molecules with noise. Molecular drug data is obtained through costly and time-consuming wet lab experiments, and this lack of data hinders the development of effective conditional generation models for drug design. Fine-tuning Methods that can achieve good performance with less data, such as LoRA\cite{hu2022lora}, may help address this problem.

\subsubsection{Flaws in Theory}
In Eq. \ref{eq:Taylor}, we treat $G$ as a continuous tensor and perform a Taylor expansion. However, the underlying molecular structure is stored discretely in $G$. This continuous approximation inevitably results in the loss of the discrete nature of the original information. Additionally, we do not have access to $\nabla_G\log\hat{q}(y_c|G^t)$ and thus make a further assumption. While the ablation experiments in Section \ref{sec:ablation} lend support to the validity of our assumption, the theoretical framework still exhibits fundamental limitations. The theoretical innovation of conditional molecule generation models is also a direction worth studying in the future.

\section{Conclusion}
In this paper, we are the first to combine conditional molecule generation models with virtual screening. With the HIV dataset, an extra classifier is trained to guide Diffusion to generate a larger proportion of HIV-inhibiting candidate molecules.  The experiment shows that our method generates more HIV-inhibiting candidate molecules than other methods. Inspired by ligand-base virtual screening, we also propose a novel metric DrugIndex which reflects the ability to generate drug-like molecules. Besides, we are the first to report the Degradation in molecule generation and find that it is from the difficulty of generating specific molecular structures. Our research brings new ideas for the application of generative models in drug design.

\section*{Acknowledgment}

This work is supported by a grant from the Natural Science Foundation of China (No. 62072070). Jiaqing Lyu is a visiting scholar from the Peking University.

\bibliography{references.bib} 

\end{document}